\documentclass[sigconf]{acmart}
\AtBeginDocument{%
  }
\usepackage{fvextra}
\usepackage{listings}
\usepackage{xcolor}

\setcopyright{acmlicensed}
\copyrightyear{2018}
\acmYear{2018}
\acmDOI{XXXXXXX.XXXXXXX}
\acmConference[Conference acronym 'XX]{Make sure to enter the correct
  conference title from your rights confirmation email}{June 03--05,
  2018}{Woodstock, NY}
\acmISBN{978-1-4503-XXXX-X/2018/06}

\begin{document}

\title{ATRIA: Adaptive Traceable ECG Reporting with Iterative Agents}

\author{Donggyun Hong, Kyuhwan Lee, Junmyung Kwon, and Yong-Yeon Jo}
\authornote{Corresponding author.}
\affiliation{%
  \institution{MedicalAI Co., Ltd.}
  \country{South Korea}
}
\email{{dghong, hwan9024, jm.kwon, yy.jo}@medicalai.com}








\renewcommand{\shortauthors}{Hong et al.}


\begin{abstract}
Existing ECG report generation is tightly coupled---interpretation and
reporting fused end-to-end, so errors propagate without stage-level
recourse---while agent-based systems decouple tasks but remain single-pass,
never revisiting earlier outputs.
Clinical ECG reporting instead unfolds iteratively, requiring progressive
context integration and bidirectional editing.
We present \textsc{ATRIA}, a multi-agent ECG reporting system that mirrors the
clinician's iterative workflow: it binds every report claim to its supporting
evidence, flags statements unsupported by that evidence, incorporates
additional context mid-session, and lets clinicians verify and revise
individual findings rather than accept one opaque output.
Because its agents use ECG analysis models already in clinical use, the
underlying findings are clinically trustworthy; and as a cloud-based web
service, \textsc{ATRIA} is ready for immediate deployment.
We demonstrate \textsc{ATRIA} through four interaction cases, 
with a live demo and video available.
\end{abstract}

\ccsdesc[500]{Computing methodologies~Neural networks}
\ccsdesc[300]{Computing methodologies~Machine learning approaches}
\ccsdesc[300]{Applied computing~Health care information systems}
\ccsdesc[100]{Information systems~Information extraction}

\keywords{ECG reporting, multi-agent systems, large language models,
clinician-in-the-loop, iterative report generation,
artifact-grounded reasoning}



\maketitle

\section{Introduction}

\textit{ECG predictive models} have advanced rapidly with deep
learning~\cite{hannun2019cardiologist, ribeiro2020automatic}, and recent work
has extended them toward LLM-based report generation through zero-shot
diagnosis~\cite{pmlr-v225-yu23b, yu2025alfred}, instruction
tuning~\cite{wan2024meit}, multimodal ECG--language
modeling~\cite{lan2025gem}, and retrieval-based
reasoning~\cite{tang2024ecgregen}.
However, these pipelines are typically \textit{tightly coupled}: ECG
interpretation and report generation are fused into a single end-to-end step,
so errors propagate to the final report without stage-level recourse.

\textit{Agent-based systems} have emerged to decouple this, each specialized
to a particular ECG interpretation task~\cite{ecgagent2026, careecg2026}.
However, their focus remains \emph{single-pass interpretation}: a task is solved
once per request, and earlier outputs are not revisited or refined.
Clinical ECG reporting instead unfolds iteratively, requiring progressive
context integration and bidirectional editing~\cite{schlapfer2017computer,
mason2007ecg_statements}---attaching lab values mid-session, comparing prior
recordings, or augmenting evidence on demand.
We therefore center our design on making ECG reporting adaptive and
traceable across the workflow, so that a clinician can verify, follow up on,
and revise each finding rather than accept or reject a single opaque output.

We present \textsc{ATRIA}---\emph{\underline{A}daptive \underline{T}raceable
\underline{R}eporting with \underline{I}terative \underline{A}gents}---a
multi-agent system that mimics the clinician's iterative ECG reporting
workflow.
Unlike single-pass ECG report generators, \textsc{ATRIA} binds every report
claim to its source evidence, verifies that each claim is supported, and lets
clinicians revise individual claims---turning one-shot generation into an
auditable, interactive workflow over ECG models already in clinical use, whose
findings are clinically trustworthy.

\textsc{ATRIA} is built around four properties: \textbf{stage-level
traceability} through inspectable stage outputs; \textbf{progressive context
integration} that accepts additional inputs (lab values, prior records)
mid-session; \textbf{bidirectional iterative use} that addresses follow-up
requests without re-executing the pipeline; and a \textbf{cloud-based,
deployable web service} integrating ECG models already in clinical
use---not a research-only prototype.
We demonstrate \textsc{ATRIA} through four interaction cases drawn from
routine ECG reading.

Our contributions are as follows:
\begin{itemize}
  \item We propose \textsc{ATRIA}, a multi-agent ECG reporting system
  that mimics the clinician's iterative reading workflow.
  \item We design \textsc{ATRIA} around coordinated stages that make ECG
  reporting traceable and iteratively revisable.
  \item We implement \textsc{ATRIA} as a cloud-based web service
  integrating ECG models already in clinical use, and demonstrate it through
  four interaction cases; a demo video and live demo are available at
  \textbf{\url{https://atria-ecg-demo.netlify.app}}.
\end{itemize}


\begin{figure}[t]
  \centering
  \setlength{\fboxsep}{0pt}%
  \setlength{\fboxrule}{0.5pt}%
  \fbox{\includegraphics[width=0.49\textwidth]{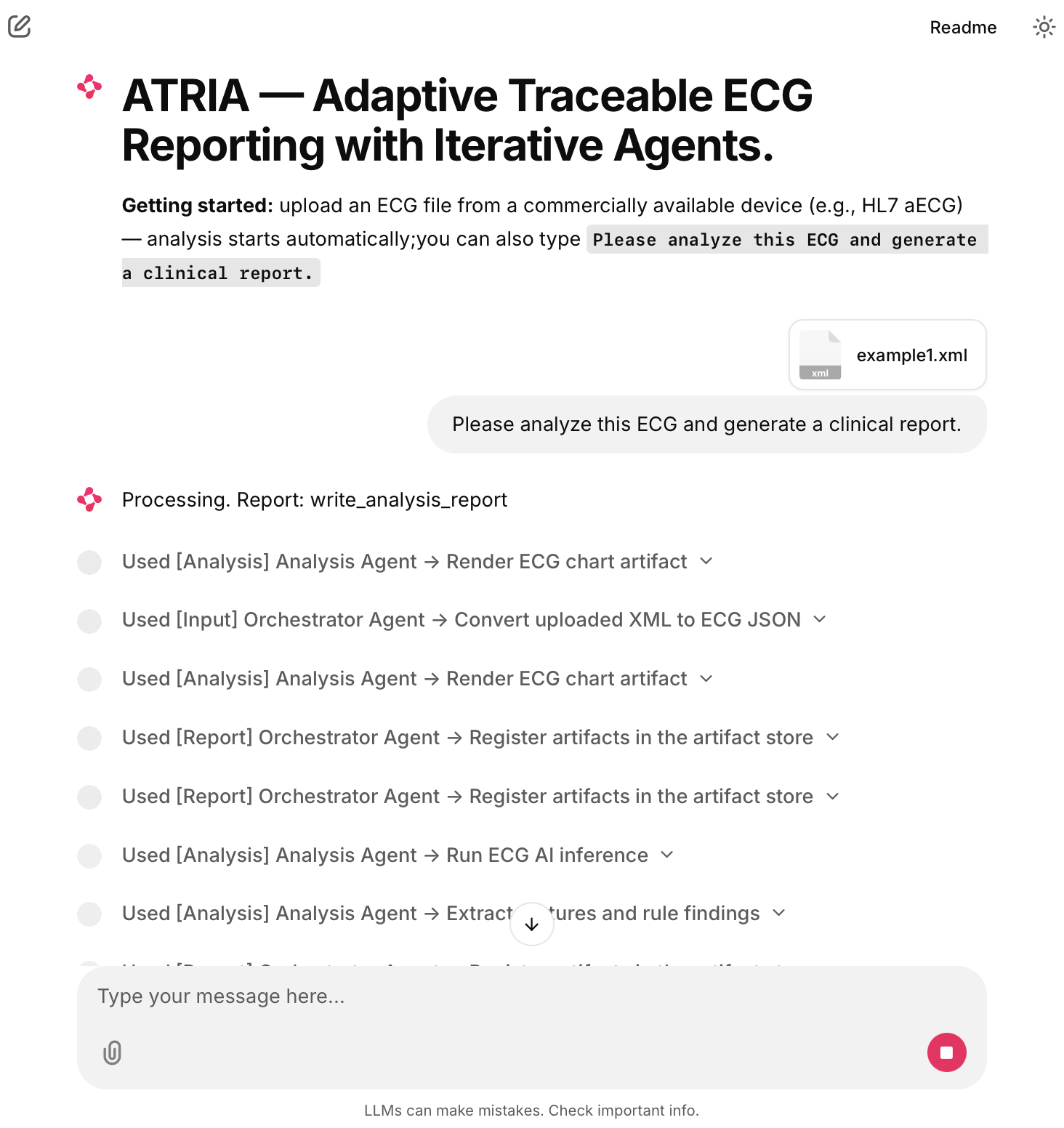}}
  \caption{Screenshot of \textsc{ATRIA} chat interface after an ECG upload: the
  \texttt{Orchestrator Agent} dispatches the staged workflow, with each agent
  step streaming in order as artifacts are produced.}
  \label{fig:demo_screen}
\end{figure}

\section{System}

\begin{figure}[t]
    \centering
    \includegraphics[width=0.49\textwidth]{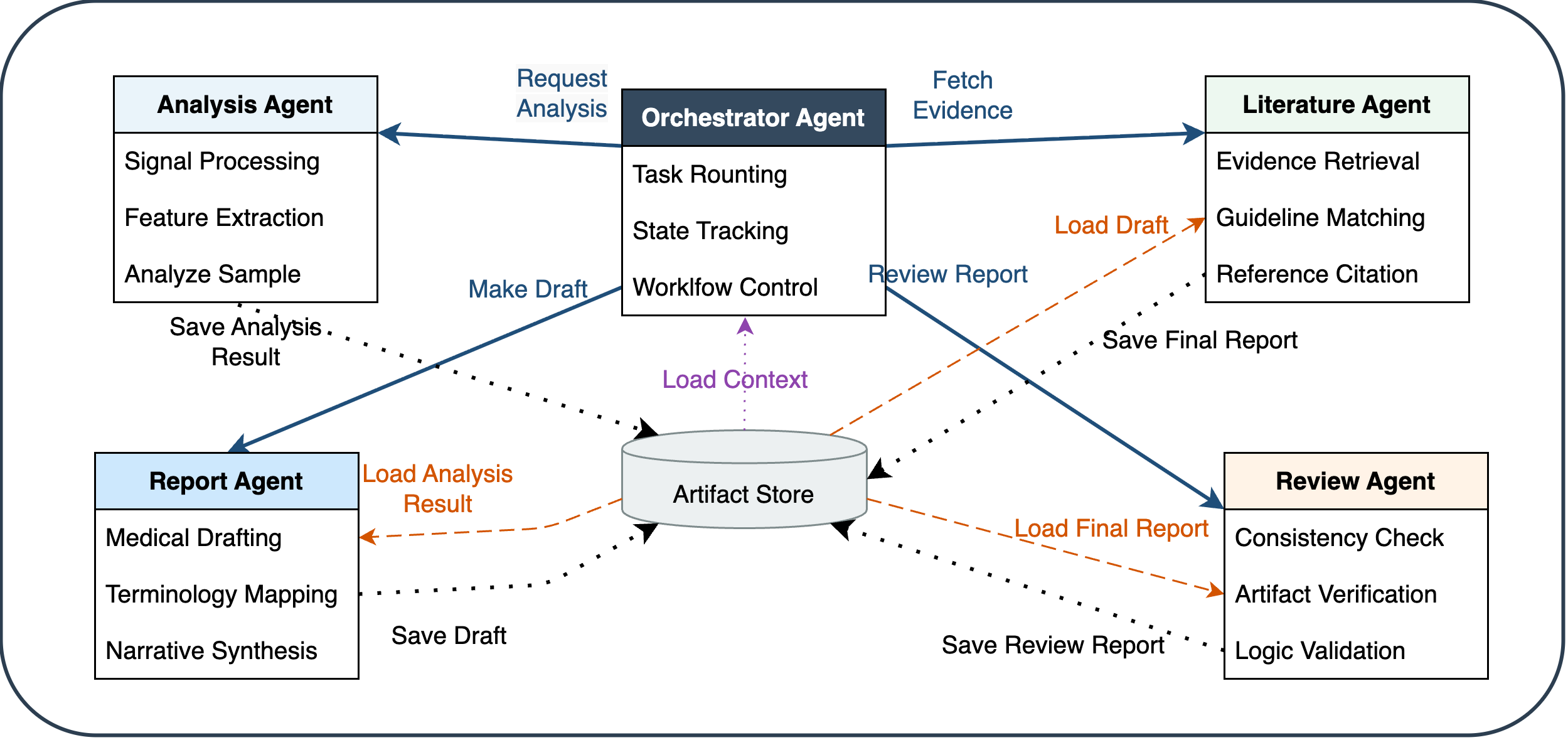}
    \caption{Overview of \textsc{ATRIA}. Five agents coordinate over a shared
    artifact store that every agent reads from and writes to, preserving
    intermediate outputs across the staged workflow.}
    \label{fig:example}
\end{figure}

\textsc{ATRIA} organizes ECG analysis and reporting into coordinated stages
whose intermediate artifacts are preserved throughout the session, supporting
iterative refinement rather than a single-pass run.
Three requirements distinguish it from single-pass pipelines---\emph{stage-level
traceability}, \emph{progressive context integration}, and \emph{bidirectional
iterative use}---and motivate two architectural decisions: stage-level handoffs
in which each stage emits an explicit, inspectable artifact rather than a fused
end-to-end output, and a shared store that every agent reads from and writes to,
so that preserved artifacts can be revisited, augmented, or selectively
re-derived without re-executing the pipeline.
Figure~\ref{fig:example} illustrates the overall framework.

\subsection{Architecture Overview}

To lower the barrier to clinical adoption, \textsc{ATRIA} exposes its workflow
through a chat-based interface---the same interaction model as common LLM-based
agent systems---so users interact by conversation rather than a specialized
tool. All language components run on \textsc{gpt-5-mini}.
A session corresponds to a single patient encounter and stays open after the
initial report, so that follow-up requests operate over the same preserved
artifacts.

The workflow proceeds in four steps.
(1)~The user submits an ECG record, optionally with additional context such as
laboratory values or prior recordings.
(2)~The \textsc{Orchestrator Agent} takes control; it performs no analysis or
generation itself and only routes work to the worker agents.
(3)~For an initial request it invokes the worker agents in sequence---analysis,
report drafting, optional literature support, and review---each writing its
output to the shared store, and the system returns a sectioned ECG report
together with supporting artifacts such as an ECG visualization and, when
requested, retrieved evidence.
(4)~For a follow-up request---clarification, context attachment, comparison,
evidence, or revision---the orchestrator dispatches only the agents whose
artifacts are affected, without re-executing the full pipeline unless the
underlying ECG record changes.

\subsection{Role of Agents}
\label{sec:agents_and_artifacts}

\paragraph{Orchestrator Agent}
It handles coordination alone---maintaining session state, routing tasks to
the appropriate worker agent, and connecting follow-up requests to preserved
artifacts---without performing analysis or report generation itself.
It may bind \emph{multiple} analysis artifacts to a single session---e.g., two
ECG recordings from the same patient---so that downstream report and review
steps can reason across records.
The worker agents write their intermediate outputs (analysis findings, drafts,
retrieved references, review feedback) to the shared store, where they remain
accessible throughout the session.

\paragraph{Analysis Agent}
It accepts an ECG record together with \emph{optional
patient context} (e.g., laboratory values such as electrolytes or cardiac
biomarkers) and produces structured findings, measurements, and signal-grounded
evidence.
It calls \emph{AiTiA}, MedicalAI's suite of ECG models in clinical
use,\footnote{MedicalAI Co., Ltd.: \url{https://medicalai.com/en/product/}} for
left ventricular systolic~\cite{rhee2026aitialvsd,
jeong2024lvsd_af_rvr} and diastolic~\cite{lim2026ecg_diastolic} dysfunction,
aortic stenosis~\cite{kwon2020aortic}, and acute myocardial
infarction~\cite{lee2025romiae},
extracts signal-level features~\cite{pmlr-v287-park25a}, and applies
rule-based summarization~\cite{ecgwaves, litfl_ecg, wikidoc_ecg}.
The analysis stage thus grounds report content in the outputs of these
production models.
Supplied laboratory values populate a separate manifest slot consumed by the
\textsc{Report Agent} so that the corresponding report sections can reference
those values.

\paragraph{Report Agent}
It consumes the analysis manifest and generates a
sectioned ECG report (Example~\ref{lst:report_example}) covering clinical
impression, feature summary, AI/rule findings, lead-wise observations,
recommendations, caveats, and attachments.
Each section is bound to the subset of upstream artifacts from which it was
derived, enabling \emph{partial re-drafting}: when only a subset of artifacts
changes---e.g., context attachment or user removal---only affected sections
are scheduled for regeneration.
When multiple analysis manifests are bound to the session, the template is
instantiated comparatively, listing the bound records section by section within
a single report rather than emitting separate per-record reports.

\begin{lstlisting}[
  caption={Template of the clinical report. 
  \href{https://atria-ecg-demo.netlify.app/sample_init_result.png}{Sample link}.},
  label={lst:report_example},
  basicstyle=\ttfamily\footnotesize,
  breaklines=true,
  columns=fullflexible,
  frame=single,
  framerule=0.5pt,
  framesep=4pt
]
## Clinical Impression    -- diagnosis, conditions
## Feature Summary        -- HR, intervals, axis, etc.
## Findings               -- AI + rule outputs
## Lead-wise Observations -- per-lead patterns
## Recommendations        -- follow-up actions
## Caveats                -- weak findings
## Attachments            -- ECG plot, evidence
\end{lstlisting}

\paragraph{Literature Agent}
It is invoked when report statements require
external grounding or when users request supporting evidence. Target claims
are selected either automatically, as statements with weak cross-artifact
support, or by a user pointing to a specific claim in a follow-up; for each, it
formulates a query and retrieves passages from an \emph{ECG-specific knowledge base
curated in-house by medical specialists}~\cite{yu2025alfred}.
Retrieved passages are stored as a literature-evidence artifact bound to the
originating statement; the agent complements but does not replace the
analysis stage.

\paragraph{Review Agent}
It examines the draft against the full set of accumulated
artifacts and checks for omitted findings, mismatches between cited evidence
and report content, and unsupported or contradictory statements---those that
reference a finding absent from the manifest or contradict the rule summary.
Structured feedback is returned to the \textsc{Report Agent} and
\textsc{Orchestrator Agent} for targeted revision and surfaced to users for
iterative review.
The agent can also be \emph{re-invoked after any artifact-changing follow-up}
(e.g., a context-augmented re-draft) to verify that the updated draft remains
consistent with the current artifact set.

\subsection{Implementation Notes}

\textsc{ATRIA} is implemented as a cloud-based web service.
The artifact store is session-keyed, so all artifacts produced within a session
remain available to later requests in that session.
Each agent uses stage-specific prompts and a common JSON schema for artifact
exchange.
ECG inputs are accepted as the raw record files exported by \emph{commercially
available ECG devices}, covering common clinical ECG formats such as HL7 aECG,
SCP-ECG, and DICOM-ECG.
Patient context (lab values, prior ECG records) is accepted as structured
uploads or inline chat attachments and populates the corresponding manifest
slots.


\section{Use Cases}
\label{sec:use_cases}

We illustrate \textsc{ATRIA}'s behavior through four cases that collectively
cover the three design requirements through distinct follow-up routing paths.
Each case corresponds to a step a clinician routinely performs when reading
and refining an ECG report---correcting unsupported statements, attaching
laboratory values, requesting supporting evidence, and comparing prior
recordings~\cite{schlapfer2017computer, mason2007ecg_statements}---rather than
capabilities chosen for the system's convenience.
The report and evidence excerpts in the following listings are condensed for
space; the actual rendered figures for each example, together with the full
interactions, are available on the demo page and in the accompanying demo
video.

\newcommand{\caseblock}[3]{%
  \noindent\textbf{Trigger.}~#1\enspace
  \textbf{Behavior.}~#2\enspace
  \textbf{Artifact.}~#3\par\smallskip%
}

\subsection{Case 1: Groundedness Verification on the Initial Report (Traceability)}

\caseblock
  {The user uploads a single ECG record and requests an initial report.}
  {The \textsc{Orchestrator Agent} coordinates the full staged workflow; the
  \textsc{Review Agent} compares the draft against the \textsc{Analysis
  Agent}'s manifest and flags a statement with no corresponding support in
  the upstream artifacts---an unsupported claim that would otherwise reach
  the final report unflagged in a single-pass pipeline.}
  {A review-feedback artifact is added to the session, flagging statements
  that lack support in the upstream artifacts so they can be removed or
  qualified before finalization (Example~\ref{lst:case1}).}
\begin{lstlisting}[
  caption={Case 1 --- Review Agent verdict on the initial draft and
  representative statements flagged as unsupported by the upstream artifacts.
  \href{https://atria-ecg-demo.netlify.app/sample_review}{Sample link}.},
  label={lst:case1},
  basicstyle=\ttfamily\footnotesize,
  breaklines=true,
  breakatwhitespace=true,
  columns=fullflexible,
  frame=single,
  framerule=0.5pt,
  framesep=4pt
]
[Review Agent verdict]
  Result: Ungrounded statements found (evidence absent). Several
  report statements are not grounded in upstream artifacts; they
  appear to be external guideline content not present in the
  supplied data.

[Ungrounded statements -- evidence absent]
  (a) "Use serial high-sensitivity troponin algorithms (0/1 h or
       0/2 h), supported by contemporary ESC and AHA/ACC guideline
       recommendations."
      -- no guideline text or troponin data in any upstream artifact.

  (b) "LVH voltage criteria (Sokolow-Lyon, Cornell) have limited
       sensitivity; correlate with echocardiography for LV mass
       assessment."
      -- rule has lvh_qrsd=1 (flag only); performance claims and the
         echo recommendation are not in any artifact.

  (c) "QTc 444 ms is in the borderline range (>=450 ms men, >=460 ms
       women; >=500 ms higher torsades risk)." [partial-support case]
      -- numeric QTc 444 supported in feature.summary; sex-specific
         cutoffs and torsades thresholds are not present.
\end{lstlisting}

\subsection{Case 2: Lab-Augmented Reporting (Progressive Context Integration)}

\caseblock
  {After an initial report is produced, the user attaches laboratory values
  (e.g., serum potassium, troponin) and asks the system to incorporate them
  into the interpretation.}
  {The \textsc{Orchestrator Agent} routes the request as a patient-context
  attachment.
  The \textsc{Analysis Agent} is re-invoked with the same ECG record but
  with the supplied laboratory values populated as a manifest slot; the
  \textsc{Report Agent} re-drafts only the affected sections (clinical
  impression, recommendations).}
  {The session's analysis manifest is augmented with the attached laboratory
  values, and the report's clinical impression is updated with a qualifier
  reflecting those values (Example~\ref{lst:case2}).}
\begin{lstlisting}[
  caption={Case 2 --- Lab payload, the lab-driven Clinical Impression revision,
  and a newly added Recommendation.
  \href{https://atria-ecg-demo.netlify.app/sample_lab.png}{Sample link}.},
  label={lst:case2},
  basicstyle=\ttfamily\footnotesize,
  breaklines=true,
  breakatwhitespace=true,
  columns=fullflexible,
  frame=single,
  framerule=0.5pt,
  framesep=4pt
]
[Lab values attached -- abnormal flags shown below]
  cardiac_biomarkers:
    troponin_I = 0.06 ng/mL  HIGH (ref<0.04),
    BNP = 155 pg/mL  HIGH (ref<100),  CK-MB = 5.4 ng/mL  BORDERLINE
  lipid_panel:
    total_chol = 224 mg/dL  HIGH,  LDL = 152 mg/dL  HIGH,
    HDL = 41 mg/dL  LOW,  triglycerides = 168 mg/dL  HIGH
  glucose_diabetes:
    fasting_glucose = 138 mg/dL  HIGH,  HbA1c = 6.9 %  DIABETES
  hematology: 
    hemoglobin = 11.8 g/dL  LOW,  hematocrit = 36 %  BORDERLINE
  electrolytes, kidney_function:  all within reference

[Clinical Impression -- before -> after]
- "no unequivocal rule-based evidence of acute ST-elevation MI;
   recommend correlation with symptoms and troponin/serial ECGs to
   exclude acute ischemia."
+ "given a mildly elevated troponin and elevated BNP, acute or 
   recent myocardial injury cannot be excluded and requires clinical
   correlation and serial testing."

[Recommendations -- newly added]
+ "Address cardiovascular risk factors: abnormal lipid profile
   (LDL 152 mg/dL, total cholesterol 224 mg/dL), and diabetes
   (HbA1c 6.9%) -- initiate or optimize risk-reduction (statin,
   glycemic control) per guidelines after acute evaluation."

\end{lstlisting}

\subsection{Case 3: Evidence Augmentation on Demand (Bidirectional Iterative Use)}

\caseblock
  {The user follows up on a specific report statement and asks for
  supporting evidence.}
  {The \textsc{Orchestrator Agent} routes the request to the \textsc{Literature
  Agent}, which formulates a query for the targeted claim and retrieves
  passages from the ECG-specific knowledge base.}
  {Retrieved passages are stored as a literature-evidence artifact bound
  to the originating statement, surfaced in the chat, and made available
  to the \textsc{Review Agent} (Example~\ref{lst:case3}).}
\begin{lstlisting}[
  caption={Case 3 --- User follow-up, retrieved passages with similarity
  scores, and the resulting evidence statement bound to the originating
  claim.
  \href{https://atria-ecg-demo.netlify.app/sample_evidence.png}{Sample link}.},
  label={lst:case3},
  basicstyle=\ttfamily\footnotesize,
  breaklines=true,
  breakatwhitespace=true,
  columns=fullflexible,
  frame=single,
  framerule=0.5pt,
  framesep=4pt
]
[User follow-up]
  "Please use the literature search tool to retrieve supporting
   evidence for the sinus bradycardia finding and enrich the report
   with the results."

[Targeted claim] (selected from report)
  "Sinus rhythm with marked bradycardia (HR ~48 bpm)."

[Literature Agent -- retrieved passages, with similarity scores]
  - LITFL (Normal Sinus Rhythm) -- sim 0.76
    "Sinus bradycardia = sinus rhythm with resting heart rate < 60 
     bpm in adults; P wave positive in I and II, inverted in aVR; 
     each P followed by a QRS; PR interval constant."
  - Braunwald (Heart Disease, ch.068) -- sim 0.80
    "Diagnosis when sinus node rate is less than ~50 bpm; P waves of
     normal contour precede each QRS; PR intv. generally constant."
  - ECGWaves (sinus-bradycardia) -- sim 0.83
    "Regular rhythm with ventricular rate slower than 50 bpm; P waves
     of constant morphology precede each QRS (P positive in lead II,
     negative in aVR)."

[literature_evidence artifact bound to claim]
  "Sinus bradycardia at 48 bpm with valid sinus P waves preceding 
   each QRS, regular R-R intervals, and stable PR intervals 
   -- consistent with the standard ECG criteria for sinus bradycardia
   [LITFL; Braunwald; ECGWaves]."
\end{lstlisting}

\subsection{Case 4: Comparative ECG Reporting (Progressive Context Integration)}

\caseblock
  {The user uploads a second ECG record from the same patient (e.g., a
  prior record without the suspected condition) and asks the system to
  compare it against the current report.}
  {The \textsc{Orchestrator Agent} invokes the \textsc{Analysis Agent} on the
  new record and binds the resulting manifest to the existing session
  alongside the first.
  The \textsc{Report Agent} re-instantiates the standard template
  \emph{comparatively}, contrasting the two records section by section.}
  {The session now holds two analysis manifests linked as a comparison pair,
  and the report contains an explicit per-section comparison
  (Example~\ref{lst:case4}).}
\begin{lstlisting}[
  caption={Case 4 --- Comparison-pair manifest and the comparative
  report instantiated section by section.
  \href{https://atria-ecg-demo.netlify.app/sample_compare.png}{Sample link}.},
  label={lst:case4},
  basicstyle=\ttfamily\footnotesize,
  breaklines=true,
  breakatwhitespace=true,
  columns=fullflexible,
  frame=single,
  framerule=0.5pt,
  framesep=4pt
]
[user follow-up]
  "Please compare these two ECG recordings and produce a per-section
   comparison."

[comparison_pair manifest bound to session]
  manifest_1 = ECG_1  (wide-complex tachycardia episode)
  manifest_2 = ECG_2  (post-conversion, AF rhythm)

[Report Agent -- comparative instantiation, section by section]

## Rhythm
  ECG 1: Rapid regular wide-complex tachycardia, most consistent 
         with ventricular tachycardia (VT).
  ECG 2: Atrial fibrillation, irregularly irregular ventricular
         response.
  Diff:  VT -> AF; major clinical change (VT immediately life
         -threatening, AF less acutely malignant but requires 
         management).

## Heart rate
  ECG 1 = 134 bpm; ECG 2 = 92 bpm
  Diff:  ventricular rate decreased substantially.

## QRS duration
  ECG 1 = 134 ms (wide); ECG 2 = 114 ms (borderline)
  Diff:  ~20 ms narrowing; supports return of native supraventricular
         conduction (consistent with true VT in ECG 1).

## Axis
  ECG 1 = -95 deg (extreme "northwest"); ECG 2 = +100 deg (right-axis)
  Diff:  large swing; ventricular focus -> native conduction.

## Findings (rule + AI)
  ECG 1: rule-VT / polymorphic VT, RVH flag, extreme axis; 
         AI LVSD/LVDD near threshold; low-probability MI.
  ECG 2: rule-AF, LPFB, QRS-based LVH, right-axis deviation; 
         AI LVSD lower probability than ECG 1; low-probability MI.
  Diff:  acute VT concern -> AF + structural conduction signals.

## Clinical interpretation
  Patient had sustained VT (ECG 1) that terminated (spontaneously 
  or by therapy) prior to ECG 2 (AF, narrower QRS). The change in 
  activation sequence supports a true ventricular rhythm in ECG 1 
  rather than SVT with aberrancy.
\end{lstlisting}

\section{Conclusion}

We presented \textsc{ATRIA}, a multi-agent ECG reporting system organized into
coordinated analysis, drafting, literature, and review stages, where follow-up
requests route back to preserved artifacts rather than re-running the pipeline.
The four cases exercise its three requirements---\emph{traceability} via
evidence-grounding verification, \emph{progressive context integration} via lab
augmentation and comparative reporting, and \emph{bidirectional iterative
use}---and \textsc{ATRIA} runs as a cloud-based, deployable web service
integrating ECG models already in clinical use, not a research-only prototype.
Future work will add richer clinical context, stronger verification, and
broader human-in-the-loop support, and extend the present demonstration toward
routine clinical use in practice.


\bibliographystyle{ACM-Reference-Format}
\bibliography{sample-base}

@article{hannun2019cardiologist,
  title     = {Cardiologist-level arrhythmia detection and classification in ambulatory electrocardiograms using a deep neural network},
  author    = {Hannun, Awni Y and Rajpurkar, Pranav and Haghpanahi, Masoumeh and Tison, Geoffrey H and Bourn, Colin and Turakhia, Mintu P and Ng, Andrew Y},
  journal   = {Nature Medicine},
  volume    = {25},
  number    = {1},
  pages     = {65--69},
  year      = {2019},
  publisher = {Nature Publishing Group}
}

@article{ribeiro2020automatic,
  title     = {Automatic diagnosis of the 12-lead ECG using a deep neural network},
  author    = {Ribeiro, Ant{\'o}nio H and Ribeiro, Manoel H and Paix{\~a}o, Gabriele M M and Oliveira, Derick M and Gomes, Paulo R and Canazart, Jos{\'e} A and Ferreira, M and Andersson, Carl R and Macfarlane, Peter W and Wagner, Patrick and others},
  journal   = {Nature Communications},
  volume    = {11},
  number    = {1},
  pages     = {1760},
  year      = {2020},
  publisher = {Nature Publishing Group}
}

@article{yu2025alfred,
  title   = {ALFRED: Ask a Large-language model For Reliable ECG Diagnosis},
  author  = {Yu, Jin and Park, JaeHo and Park, TaeJun and Kim, Gyurin and Lee, JiHyun and Lee, Min Sung and Kwon, Joon-myoung and Son, Jeong Min and Jo, Yong-Yeon},
  journal = {arXiv preprint arXiv:2505.03781},
  year    = {2025}
}

@inproceedings{pmlr-v225-yu23b,
  title     = {Zero-Shot ECG Diagnosis with Large Language Models and Retrieval-Augmented Generation},
  author    = {Yu, Han and Guo, Peikun and Sano, Akane},
  booktitle = {Proceedings of the 3rd Machine Learning for Health Symposium},
  pages     = {650--663},
  volume    = {225},
  series    = {Proceedings of Machine Learning Research},
  year      = {2023},
  publisher = {PMLR},
  url       = {https://proceedings.mlr.press/v225/yu23b.html}
}

@article{wan2024meit,
  title   = {MEIT: Multi-Modal Electrocardiogram Instruction Tuning on Large Language Models for Report Generation},
  author  = {Wan, Zhongwei and Liu, Che and Wang, Xin and Tao, Chaofan and Shen, Hui and Peng, Zhenwu and Fu, Jie and Arcucci, Rossella and Yao, Huaxiu and Zhang, Mi},
  journal = {arXiv preprint arXiv:2403.04945},
  year    = {2024},
  url     = {https://arxiv.org/abs/2403.04945}
}

@misc{tang2024ecgregen,
  title         = {Electrocardiogram Report Generation and Question Answering via Retrieval-Augmented Self-Supervised Modeling},
  author        = {Tang, Jialu and Xia, Tong and Lu, Yuan and Mascolo, Cecilia and Saeed, Aaqib},
  year          = {2024},
  eprint        = {2409.08788},
  archiveprefix = {arXiv},
  primaryclass  = {cs.LG},
  url           = {https://arxiv.org/abs/2409.08788}
}

@article{lan2025gem,
  title   = {Gem: Empowering mllm for grounded ecg understanding with time series and images},
  author  = {Lan, Xiang and Wu, Feng and He, Kai and Zhao, Qinghao and Hong, Shenda and Feng, Mengling},
  journal = {arXiv preprint arXiv:2503.06073},
  year    = {2025}
}

@article{schlapfer2017computer,
  title   = {Computer-Interpreted Electrocardiograms: Benefits and Limitations},
  author  = {Schl{\"a}pfer, J{\"u}rg and Wellens, Hein J.},
  journal = {Journal of the American College of Cardiology},
  year    = {2017},
  volume  = {70},
  number  = {9},
  pages   = {1183--1192},
  doi     = {10.1016/j.jacc.2017.07.723},
  url     = {https://www.jacc.org/doi/abs/10.1016/j.jacc.2017.07.723}
}

@article{mason2007ecg_statements,
  title   = {Recommendations for the Standardization and Interpretation of the Electrocardiogram: Part II: Electrocardiography Diagnostic Statement List},
  author  = {Mason, J. Warren and Hancock, E. William and Gettes, Leonard S. and Bailey, James J. and Childers, Rory and Deal, Barbara J. and Josephson, Mark and Kligfield, Paul and Kors, Jan A. and Macfarlane, Peter and Pahlm, Olle and Mirvis, David M. and Okin, Peter and Rautaharju, Pentti and van Herpen, Gerard and Wagner, Galen S. and Wellens, Hein},
  journal = {Circulation},
  year    = {2007},
  volume  = {115},
  number  = {10},
  pages   = {1325--1332},
  doi     = {10.1161/CIRCULATIONAHA.106.180201},
  url     = {https://www.ahajournals.org/doi/10.1161/circulationaha.106.180201}
}

@inproceedings{pmlr-v287-park25a,
  title     = {Benchmarking ECG Delineation using Deep Neural Network-based Semantic Segmentation Models},
  author    = {Park, Jaeho and Park, TaeJun and Kwon, Joon-myoung and Jo, Yong-Yeon},
  booktitle = {Proceedings of the sixth Conference on Health, Inference, and Learning},
  pages     = {63--88},
  year      = {2025},
  editor    = {Xu, Xuhai Orson and Choi, Edward and Singhal, Pankhuri and Gerych, Walter and Tang, Shengpu and Agrawal, Monica and Subbaswamy, Adarsh and Sizikova, Elena and Dunn, Jessilyn and Daneshjou, Roxana and Sarker, Tasmie and McDermott, Matthew and Chen, Irene},
  volume    = {287},
  series    = {Proceedings of Machine Learning Research},
  month     = {25--27 Jun},
  publisher = {PMLR},
  url       = {https://proceedings.mlr.press/v287/park25a.html}
}

@article{ecgagent2026,
  title   = {{ECG-Agent}: On-Device Tool-Calling Agent for {ECG} Multi-Turn Dialogue},
  author  = {Chung, Hyunseung and Oh, Jungwoo and Kyung, Daeun and Kim, Jiho and Kwon, Yeonsu and Kim, Min-Gyu and Choi, Edward},
  journal = {IEEE Xplore},
  year    = {2026},
  url     = {https://ieeexplore.ieee.org/document/11464123},
  note    = {IEEE Xplore document 11464123; corresponding arXiv preprint: arXiv:2601.20323}
}

@article{lim2026ecg_diastolic,
  title   = {Artificial Intelligence-Enabled {ECG} for Elevated {E/e'} on Echocardiography: Hemodynamic Relevance and Prognostic Value},
  author  = {Lim, Jaehyun and Lee, Min Sung and Suh, Jung Ho and Kang, Sora and Lee, Hak Seung and Jang, Jong-Hwan and Son, Jeong Min and Kwon, Joon-Myoung and Kim, Yong-Jin and Kim, Kyung-Hee and Lee, Seung-Pyo},
  journal = {Journal of the American Heart Association},
  volume  = {15},
  number  = {9},
  pages   = {e046989},
  year    = {2026},
  doi     = {10.1161/JAHA.125.046989}
}

@article{kwon2020aortic,
  title   = {Deep Learning-Based Algorithm for Detecting Aortic Stenosis Using Electrocardiography},
  author  = {Kwon, Joon-Myoung and Lee, Soo Youn and Jeon, Ki-Hyun and Lee, Yeha and Kim, Kyung-Hee and Park, Jinsik and Oh, Byung-Hee and Lee, Myong-Mook},
  journal = {Journal of the American Heart Association},
  volume  = {9},
  number  = {7},
  pages   = {e014717},
  year    = {2020},
  doi     = {10.1161/JAHA.119.014717}
}

@article{rhee2026aitialvsd,
  title   = {Artificial Intelligence-Driven Electrocardiogram Screening for Asymptomatic Left Ventricular Systolic Dysfunction in the General Population},
  author  = {Rhee, Tae-Min and Kang, Sora and Lee, Min Sung and Han, Ga In and Yoo, Ah-Hyun and Jang, Jong-Hwan and Jo, Yong-Yeon and Son, Jeong Min and Kwon, Joon-Myoung and Choi, Su-Yeon and Lee, Hak Seung and Lee, Heesun},
  journal = {JACC: Advances},
  volume  = {5},
  number  = {4},
  pages   = {102660},
  year    = {2026},
  doi     = {10.1016/j.jacadv.2026.102660}
}

@article{lee2025romiae,
  title   = {Artificial intelligence applied to electrocardiogram to rule out acute myocardial infarction: the {ROMIAE} multicentre study},
  author  = {Lee, Min Sung and Shin, Tae Gun and Lee, Youngjoo and Kim, Dong Hoon and Choi, Sung Hyuk and Cho, Hanjin and Lee, Mi Jin and Jeong, Ki Young and Kim, Won Young and Min, Young Gi and Han, Chul and Yoon, Jae Chol and Jung, Eujene and Kim, Woo Jeong and Ahn, Chiwon and Seo, Jeong Yeol and Lim, Tae Ho and Kim, Jae Seong and Choi, Jeff and Kwon, Joon-Myoung and Kim, Kyuseok},
  journal = {European Heart Journal},
  volume  = {46},
  number  = {20},
  pages   = {1917--1929},
  year    = {2025},
  doi     = {10.1093/eurheartj/ehaf004}
}

@article{jeong2024lvsd_af_rvr,
  title   = {Deep learning algorithm for predicting left ventricular systolic dysfunction in atrial fibrillation with rapid ventricular response},
  author  = {Jeong, Joo Hee and Kang, Sora and Lee, Hak Seung and Lee, Min Sung and Son, Jeong Min and Kwon, Joon-Myung and Lee, Hyoung Seok and Choi, Yun Young and Kim, So Ree and Cho, Dong-Hyuk and Kim, Yun Gi and Kim, Mi-Na and Shim, Jaemin and Park, Seong-Mi and Kim, Young-Hoon and Choi, Jong-Il},
  journal = {European Heart Journal -- Digital Health},
  volume  = {5},
  number  = {6},
  pages   = {683--691},
  year    = {2024},
  doi     = {10.1093/ehjdh/ztae062}
}

@misc{careecg2026,
  title         = {{CARE-ECG}: Causal Agent-based Reasoning for Explainable and Counterfactual ECG Interpretation},
  author        = {{UC Irvine and ASU and CSU authors}},
  year          = {2026},
  eprint        = {2604.10420},
  archivePrefix = {arXiv},
  primaryClass  = {cs.LG},
  url           = {https://arxiv.org/abs/2604.10420}
}

@misc{ecgwaves,
  title        = {The {ECG} Book},
  author       = {{ECGwaves}},
  howpublished = {\url{https://ecgwaves.com/course/the-ecg-book/}},
  note         = {Accessed: 2026-06-02}
}

@misc{litfl_ecg,
  title        = {{ECG} Library},
  author       = {{Life in the Fast Lane}},
  howpublished = {\url{https://litfl.com/ecg-library/}},
  note         = {Accessed: 2026-06-02}
}

@misc{wikidoc_ecg,
  title        = {{ECG} Criteria},
  author       = {{WikiDoc}},
  howpublished = {\url{https://www.wikidoc.org/index.php/ECG_Criteria}},
  note         = {Accessed: 2026-06-02}
}


\end{document}